\documentclass{article}

\usepackage[final,nonatbib]{neurips_2021}

\usepackage{comment}

\usepackage[utf8]{inputenc} % allow utf-8 input
\usepackage[T1]{fontenc}    % use 8-bit T1 fonts
\usepackage{hyperref}       % hyperlinks
\usepackage{url}            % simple URL typesetting
\usepackage{booktabs}       % professional-quality tables
\usepackage{amsfonts}       % blackboard math symbols
\usepackage{nicefrac}       % compact symbols for 1/2, etc.
\usepackage{microtype}      % microtypography
\usepackage{xcolor}         % colors

\usepackage{subfig}
\usepackage{graphicx}
\newcommand{\vp}[1]{{\color{purple}[\textbf{VP}: #1]}}
\newcommand{\vinod}[1]{{\color{purple}[\textbf{VP}: #1]}}
\newcommand{\ben}[1]{{\color{blue}[\textbf{BH}: #1]}}
\newcommand{\negar}[1]{{\color{magenta}[\textbf{NR}: #1]}}

\newcommand{\christina}[1]{{\color{teal}[\textbf{CG}: #1]}}

\title{Thinking Beyond Distributions in Testing\\ Machine Learned Models}
\author{%
  Negar Rostamzadeh \\
  Google Research\\
  \texttt{nrostamzadeh@google.com} \\
  \And
  Ben Hutchinson
 \\
   Google Research \\
   \texttt{benhutch@google.com} \\
  \AND
   Christina Greer
 \\
  Google Research \\
   \texttt{ckuhn@google.com} \\
  \And
  Vinodkumar Prabhakaran \\
  Google Research \\
  \texttt{vinodkpg@google.com} \\
}

\begin{document}

\maketitle

\begin{abstract}
  Testing practices within the machine learning (ML) community have centered around assessing a learned model's predictive performance measured against a test dataset, 
  often drawn from the same distribution as the training dataset.
%   . This test dataset is often drawn from the same distribution as the dataset used to train the model, and hence is expected to follow the same distribution as the training dataset. 
  While recent work on robustness and fairness testing within the ML community has pointed to the importance of testing against distributional shifts, these efforts also focus on estimating the likelihood of the model making an error against a reference dataset/distribution.
  We argue that this view of testing actively discourages researchers and developers from looking into other sources of robustness failures, for instance corner cases which may have severe undesirable impacts. We draw parallels with decades of work within software engineering testing focused on assessing a software system against various stress conditions, including corner cases, as opposed to solely focusing on average-case behaviour. Finally, we put forth a set of recommendations to broaden the view of machine learning testing to a rigorous practice.
\end{abstract}

\section{Introduction}
In Machine Learning (ML), \emph{testing} often refers to the evaluation of a trained model on an unseen held-out test dataset, often expected to follow the same distribution as the training data.
While such an evaluation shows the researcher (and the consumer of the research) how well the training algorithm captures phenomena of interest from the training data, and generalizes to unseen data from the same distribution, it does not provide any guarantees around a model's behavior when it is deployed in practice \cite{d2020underspecification}.
In other words, such testing is meant to empirically evaluate the efficacy of certain sources of data, training algorithms, feature(s), or representational choices, rather than to estimate the utility and harms of the system when deployed on real world data \cite{d2020underspecification,singla2021understanding}.  
% This is not a new concern \cite{d2020underspecification,singla2021understanding}. 

Prior work addresses this shortcoming in testing by evaluating the performance of models across domains \cite{hendrycks2021natural,zajac2019adversarial}, introducing domain adaptation techniques \cite{ganin2015unsupervised}, measuring performance under distribution shift \cite{rabanser2018failing}, or collecting datasets that reflect target domain distributions or characteristics \cite{dvijotham2020framework,hoffman2018cycada,hendrycks2019benchmarking}. However, these efforts tie \textit{testing an ML model} to testing against a dataset with a distribution that matches a target domain \cite{chasalow2021representativeness}.
% ---what we call ``target-distribution'' \cite{chasalow2021representativeness}.
% 
We argue that such distribution-driven testing
implicitly makes assumptions about the costs of failures which often underestimate the importance and severity of failures in the tail of the distribution.
% \BHedit{implicitly makes assumptions about the costs of failures which often underestimate the importance and severity of failures in the tail of the distribution}. 
% \BHcomment{I prefer leaning into this underestimation framing rather than the over-indexing one... wdyt?}
% \BHdelete{over-indexes on  
% the most common types of examples encountered---or the head of the distribution---relegating less likely events to a small part of the overall loss or error rate, implicitly downplaying their impact. }
However, it is often the ``black swan'' events in the tail of the distribution that lead to nonlinearities in behavior that result in unsafe outcomes (e.g., the meltdown security vulnerability in hardware \cite{lipp2018meltdown,hua2018epti}).
Even an ML system with $99\%$ accuracy may have severe vulnerabilities that are masked within the $1\%$ error rate. Furthermore, traditional ML testing does not account for the different severities of harms different errors cause in deployment, or whether they affect different subgroups of people at different rates \cite{buolamwini2018gender}.

In this paper, we provide a broad account of the various such methodological shortcomings in data distribution driven machine learning testing. We then draw on testing practices from other mature engineering disciplines and provide a set of comprehensive recommendations towards rigorous ML testing, along the conditions to test for, the processes that guide this testing process, and the artefacts that can be a step towards ML system reliability.

\section{Methodological Shortcomings in ML Model Testing}
\label{ml_testing}
% \section{\BHedit{Methodological Shortcomings in ML Model Testing} \BHdelete{Missing Considerations in Machine Learning Testing}}\label{ml_testing}
% \BHcomment{I didn't feel ``missing considerations'' was accurately reflecting the bolded paragraph titles below. wdyt? feel free to revert of course!}
% \VPcomment{Looks good to me.}

% \begin{figure}%
%     \centering
%     \subfloat[\centering Typical coverage in SW testing]{{\includegraphics[width=.45\textwidth]{AuthorKit20/fig1.png} }}%
%     \qquad
%     \subfloat[\centering Typical coverage in ML testing]{{\includegraphics[width=.45\textwidth]{AuthorKit20/fig2.png} }}%
%     \caption{Comparison of Testing Coverage Scenarios.}%
%     \label{fig_graph}%
% \end{figure}
Historically, testing of AI systems was not exclusively statistical in nature \cite{jones1995evaluating}. For instance, \cite{guida1986evaluation} lays out how one might construct test sets for evaluating specific aspects of a Natural Language Processing (NLP) system by using both a randomly sampled dataset to evaluate the conceptual competence of the system, and a dataset of instances containing different linguistic variations of the same sentence conveying the same meaning, as a way to evaluate the linguistic competence of the NLP system. The switch from such a broad view of what evaluation means, to a singular focus on statistical evaluations based on held-out test sets is arguably a side effect of the overwhelming success of statistical NLP 
% \BHcomment{what is non-statistical ML?} 
in the 90s. 
From a system reliability perspective, the goal of model testing should ideally include considerations such as a) robustness to distribution shifts, b) distributions of errors, and c) severities of those errors.
These factors are often overlooked, or relegated to optional, auxiliary testing efforts in much of ML research and development.  
% However, this over-reliance on statistical testing is not really warranted from a system reliability perspective \BHcomment{I feel like this point needs more setting up. e.g. one argument could be that the goal of model testing is deep understanding of model utility, which might entail knowing various things such as a) robustness to distribution shifts, b) distributions of errors, and c) severities of errors}, and results in overlooking numerous failings along robustness, fairness, and other harmful system behaviours. 
In this section, we outline six core shortcomings that results from the common practices within machine learning testing. 
These six shortcomings are not meant to be mutually exclusive categories; rather, each building on the ones discussed prior to them. 
% In doing this, we will draw parallels from other testing paradigms, especially software testing. 

\textbf{Treating all examples as equal}: The dominant paradigm of measuring model accuracy on an \textit{unseen} i.i.d.\ dataset drawn from the same distribution as the training data mitigates against the pitfalls of over-fitting during learning. However it also makes two critical assumptions which reduce the diagnostic power of the testing. 
% \BHcomment{I'm experimenting with this phrase ``diagnostic power'', analogous to the concept of power in null hypothesis testing. wdyt??}
Firstly, in weighting all data points in the test set equally, the measurement is most sensitive to failures in model performance on the head of the distribution---and conversely least sensitive to failures in model performance on the tail. For example, if an image understanding model is evaluated on a dataset in which humans appear in only a small fraction of images then those images, and the errors the model makes on images with humans, will be given little weight in the evaluation.  
Secondly, the comparison of predictions (i.e., model outputs) with expectations ignores any available context. Notably, it ignores features of the model inputs. For example, the evaluation of an image analysis model for detecting stop signs would ignore the important factor of whether there are also pedestrians or cyclists in the image.  
%  \BHcomment{I took a stab at reframing the ``misalignment of goals'' point to be about the simplification of treating all examples as equal. wdyt? feel free to revert if you prefer the former!}

% \BHdelete{
% \textbf{Misalignment of goals}: Arguably, statistical testing is a compelling step to empirically establish the general net impact of a new source of data, new algorithm, features, or representations. Compared to other disciplines that sometimes measure the \textit{fit} of a particular model against the entire dataset to make claims about patterns within that dataset, the practice of measuring performance against a held-out dataset is a superior evaluation technique to avoid pitfalls such as over-fitting, for instance \BHcomment{This feels a bit like a strawperson argument. But I do think there is an interesting point to be made that ML does in fact sometimes compare performance on training data and held out data in order to measure overfitting.}. However, it is important to distinguish this scientific goal of average-case behaviour \BHcomment{I would typically characterise scientific goals as prototypically involving hypothesis testing and/or seeking explanations... Is there an alternative way to discuss the goal of average-case testing? } from an engineering goal of testing when appropriate. The lack of recognition of this misalignment in goals is one of the core reasons for the overemphasis on statistical testing as the dominant approach across board in ML applications as well.}

% \BHedit{
\textbf{Treating all failures as equal}: Suppose a computer vision system misidentifies a person as an animal. This is a significantly different type of error than, say, misidentifying a cat for a dog. The dominant ML testing paradigm fails to account for the fact that not all errors are qualitatively equal, and thus ignores dignitary and other harms. This simplification reduces the labour and other efficiency costs involved in doing contextualized evaluations of classes of errors. We similarly see these economic imperatives at play in the focus of much of the work on (un)fairness in NLP being focused on gender bias specifically---compared to other axes of biases---because gender-labeled datasets are relatively easier to assemble \cite{buolamwini2018gender}.
Similarly, most of the testing efforts outlined above treat models in isolation, considering just the inputs and outputs of the model, whereas, it is important to consider how the model integrates into the environment where it will be introduced, accounting for delayed impacts and feedback loops \cite{martin2020extending}. 
% }  \BHcomment{I took a stab at reframing the ``overemphasis on empiricism'' point to be about the severity of the errors. wdyt? (I'm not sure how the point about economics quite fits in *sad face* but these points were adjacent in the previous version too.) feel free to revert if you prefer the former!}

% \BHdelete{
% \textbf{Overemphasis on empiricism}: The  overwhelming success of statistical methods in ML modeling has also contributed to an overemphasis on statistical methods for evaluation. For instance, when faced with fairness/robustness failures, most efforts to measure and mitigate them tend to focus on cases where one can reliably and easily collect datasets with additional information that allows such measurement. This is reflected in how most work on (un)fairness in NLP focus on gender bias, compared to other axes of injustices, because gender-labeled datasets are relatively easier to assemble \cite{}. While such statistical evaluations of fairness/robustness failures are crucial, this overemphasis fails to account for the fact that not all errors are equal, and obscures system failures that are maybe more meaningful in real world scenarios, but less amenable to empirical evaluations. For instance, a computer vision system misidentifying a person as an animal, especially if it reinforces decades worth of harmful stereotypes, is a significantly different error than, say, misidentifying a cat for a dog. }

\textbf{Overlooking corner cases}: As outlined above, the traditional ML testing often focuses on average-case behaviour of the model. Each data point in a test set can be viewed as an individual test, and whether or not the data point, or test, passes or fails makes up part of the accuracy, which is then reported as a metric for system performance. However, this approach often does not look into which data points are failing the test, or whether there is a pattern there. Often times the errors correspond to the edge cases that are a minority, while the performance metric focuses on the head of the distribution. Recent work pointing to fairness and robustness failures has prompted work on teasing apart the evaluation set, and associating a subset of that set with a behavior one wants to test for. For instance, disaggregated fairness evaluation \cite{mitchell2019model} is essentially about performing the test on subsets of the data that have a certain feature held constant. Similarly, robustness testing often involves examining the tail distribution for certain edge cases--things like natural adversarial examples \cite{nie2020adversarial,hendrycks2021natural}, or typos for NLP \cite{checklist:acl20}. In other words, defining metrics for fairness and robustness can in essence be thought of as adding specific attention to sections of the tail distribution.
% \vp{Close this down re-stating the problem.}
% \BHcomment{I would like to suggest making this a bit more general than ``corner cases'' as that very phrase already assumes that the input space has ``corners'' an/or edges. Instead, I think this point might 1) be re-titled something like ``Ignoring the topology of the inputs'', and 2) argue for the relevance of notions such as ``edge cases'', ``corner cases'' and ``similar cases'', all of which require some form of topology.}

% \textbf{Isolated model testing}:  Most of the testing efforts outlined above treat models in isolation, considering just the inputs and outputs of the model. However, it is important to consider how the model integrates into the environment where it will be introduced, accounting for delayed impacts and feedback loops \cite{martin2020extending}. \vp{Expand this.}
% \BHcomment{I wonder if this point might be moved up to be closer to the point about how misclassifying people as animals is worse than classifying cats as dogs?}\negar{I agree that we could take this point exactly under the first two points}

\textbf{Lack of an integrated process}: 
It is important to note that recent research has begun inquiries into ML testing for the edge cases. For instance, fairness testing or disaggregated testing \cite{buolamwini2018gender,mitchell2019model} separately assesses model performance against data pertaining to different subgroups of people; counterfactual testing \cite{kusner2017counterfactual} such as perturbation sensitivity analysis \cite{prabhakaran2019perturbation} assesses model behaviour in response to small controlled changes in the input, similar to unit testing. Approaches such as the fairness gym \cite{d2020fairness} or system dynamics based simulations \cite{martin2020extending} can be thought of as instances of integration testing in simulated environments. However, these efforts largely continue to be disconnected efforts with those specific robustness goals, rather than being integrated into a unified or standard process of ML development, even for critical production systems.

% While recent years have seen more work on fairness and robustness testing, such as disaggregated evaluation \cite{}, using synthetic examples in the testing data \cite{}, counterfactual analysis \cite{}, sensitivity analysis \cite{} etc., these testing efforts are often closely integrated into the standard process of ML development even for critical production systems in the same way, as a held-out test performance measurement is. Such an integration is crucial because \vp{we say so}.
% \christina{Should we also touch on how such practices are considered, but not integrated into the standard process of ML development even for critical production systems in the same way that test practices for other production software is?}

\textbf{Lack of artifacts for transparency}: One of the consequences of not having integrated processes around comprehensive testing in ML is that there is no standard mechanism to communicate which tests have been performed on the ML model, and which ones of them succeeded vs. failed. While frameworks such as model cards \cite{mitchell2019model} provide a great transparency mechanism, most implemented model cards in practice reports largely on traditional held-out testing results, and sometimes disaggregated fairness evaluation. Lack of such standardized mechanisms to communicate the test results of a comprehensive suite of test cases puts the end users and stakeholders at a disadvantage. 

\section{Towards More Comprehensive Testing in ML}\label{swe_practices} 
\begin{comment}
\negar{I feel like informative may be interpreted as information definition in information theory but I like replacing rigorous testing with something that has more weight on the information.}
\BHcomment{I think leaning into informativity might be nicer than implying previously testing wasn't rigorous, wdyt?}

\end{comment}

Establishing guarantees of an ML model's behaviour in real-world contexts should become a core criteria for its adoption in real-world application scenarios. In this section, we 
% re-imagine what testing means in the context of machine learning research and development, and
put forth a set of recommendations towards rigorous testing practices that can provide strong, reliable, and transparent guarantees for ML system reliability. Our recommendations pertain to practices along three layers: what \textit{conditions} are tested for, what \textit{processes} guide the testing, and what \textit{artifacts} communicate the test results to the model consumers. 
These recommendations are not meant as a set of concrete steps for practitioners, rather a set of considerations that should shape the steps towards comprehensive testing that is appropriate for the application context.
We draw inspiration from mature testing practices in different engineering disciplines, especially software engineering, as well as testing-oriented engineering paradigms such as test-driven development. Our recommendations are aimed at applications where guarantees on real-world performance and reliability are crucial; however, these recommendations are also relevant in research scenarios in high-stakes domains, such as health care. 

\subsection{\textit{Conditions} that are tested for}

As outlined in Section~\ref{ml_testing}, current ML testing predominantly focuses on measuring average case behaviour of models, in isolated test environments, to statistically establish superiority of the data and/or algorithms used to build the models. This averaging is done in at least three ways, each of which reduces the informativeness of the test results: i) all data points are weighted equally, ii) all contexts are treated as equal, and iii) all error types are weighted equally.
We argue that in real-world applications, one must shift the focus to an engineering-motivated goal of ensuring that the model will work as desired in the contexts it will be deployed in. While testing against a target distribution is an effort in this direction, it does not address all the shortcomings. In software testing, for instance, unit testing is a step used to isolate and examine a small piece of code against potential failure modes (or test cases) that are designed by the developer/tester who knows the expected behaviour of that module. 
% It involves testing for edge cases and potential failure modes, aimed at improving reliability and robustness of software systems. 
What could unit testing look like in the case of ML testing? Recent research on designing test sets that focus on failure cases close to natural data distribution \cite{hendrycks2021natural}, introducing adversarial examples for robustness \cite{nie2020adversarial,goel-etal-2021-robustness}, and behavioral testing of NLP models \cite{checklist:acl20} can all be considered as instances of unit testing. 
% The idea of behavioral testing of NLP models using checklists also evokes the analogy with unit testing \cite{checklist:acl20}. 
Similarly, it is important to test for emergent failure modes when different component ML models are integrated as a part of a larger system, since the compounding of errors may follow patterns undetectable during isolated testing \cite{martin2020extending}. 
% Many computer vision applications are addressed by an ensemble of multiple modules, and it is also very important to check if those corner cases are covered when modules are integrated as a part of a larger system / application. This practice is called integration testing in S/W testing development.
In addition, failure cases are not all similar and from the same priority. For example misclassifying a can on the street can be an edge case of a self-driving car, but its severity is not as great as misclassifying a pedestrian and traffic sign in the same scenario. These two examples are not of the same severity and shouldn't be treated as such.

\vspace{20pt}

\textit{Recommendations:}
\begin{itemize}
    \item Consider going beyond the average case behavior of the system, and design test cases that cover corner cases of interest specific to the task.
    % that could be costly and harmful if they happen (even rarely).
    % Design tests that are covering edge cases in an ML system.
    \item Consider the context in which the system will be deployed and assign the severity of failing the tests, in order to prioritize testing and subsequent fixes.
    % \item Consider the context and task in place when prioritizing unit tests. 
    \item Consider test cases that account for the societal disparities across different subgroups and how failures might impact different subgroups in different rates.
    \item Make sure that test cases are also designed for scenarios where the ML model is integrated with other ML and non-ML components.
    % your application is a combination of multiple modules and functions. For example if your background subtraction model, has a face detector as part of the system, make sure not only the face detector works well alone, but it also works as expected.
\end{itemize}

\subsection{\textit{Processes} that guide the testing}
% Statistically speaking, t
In the traditional software development process, testing takes up significant amount of time, resources, and effort \cite{harrold2000testing}. Even moderate-sized software projects takes up hundreds of person-hours dedicated to writing test cases --- conditions to test for and pass/fail criteria, implementing them, and meticulously documenting the results of those tests. In fact, software testing is considered an art \cite{myers2011art} requiring its own technical and non-technical skills \cite{sanchez2020beyond,matturro2013soft}, and entire career paths are built around testing \cite{cunningham2019software}. 
\textit{Test-driven development}, often associated with agile software engineering frameworks, is a practice which integrates testing considerations in all parts of the development process \cite{astels2003test,george2004structured}. These processes rely on a deep understanding of the software requirements as well as user behavior modeling, in order to anticipate failure modes during deployment, as well as continued expansion of the test suite. In contrast, ML testing is often relegated to a small portion of the ML development process, and predominantly focuses on a static snapshot of data to provide performance guarantees. Despite the growing research into fairness and robustness failures in ML models, these efforts are often relegated to auxiliary steps that are not fully integrated into a typical ML development process. Furthermore, identifying failure modes in a deployed ML system is not always straightforward; it require a deep understanding of the societal ecosystems surrounding ML interventions \cite{sambasivan2021re}.

\textit{Recommendations:}

\begin{itemize}
    \item Consider anticipating, planning for, and integrating testing in all stages of development process, research problem ideation, the setting of objectives, and system implementation.
    \item Consider building a practice around documenting desirable behaviours of the ML system as test cases, and how to bring diverse perspectives into designing this test suite. 
    % spending a significant portion of time and resources on testing. Robustness testing shouldn't be an optional consideration. Consider integrating it in your test case design. 
    \item Consider participatory approaches (e.g., \cite{martin2020extending}) to ensure that the test suite accounts for the societal contexts and the embedded values within which the ML system will be deployed. 
    % Plan for proactively learn about the broader social impact of technologies that you are creating and values of the society as a first step in designing your system test case design.
\end{itemize}
\begin{comment}

\textbf{Test-driven development} is another practice that is been implemented in agile software engineering communities, which embed and think about testing design, prior to developing the framework. This practice is not only helping with evaluations, but help with thinking about the correct behavior of the system as well as what could go wrong during the development process. In the ML community, methodologies like model card \cite{}, encourage planning and transparency in all steps of ml system development process. A clear point in an ML system, it could be very hard if not impossible to plan for developing a system that works for all edge cases. What matters is the priority. What are the situations that can happen that could specifically harm a subgroup? What are the potential cases that caused substantial harms in the past? 
\end{comment}

\subsection{\textit{Artifacts} that document the test results}

Recent work has pointed to the importance of standard frameworks for transparency \cite{gebru2018datasheets,mitchell2019model} and accountability \cite{raji2020closing} in ML based interventions. While transparency artifacts such as datasheets for datasets \cite{gebru2018datasheets} and model cards \cite{mitchell2019model} are frameworks derived from practices in other mature domains such as hardware documentation practices or food nutrition labeling, these frameworks currently do not provide a way to communicate comprehensive test results. As \cite{harrold2000testing} points out, software testing produces a number of artifacts including execution traces, test results, as well as test coverage information. In addition to increased transparency, such information serves also as a way to iteratively build new test cases for newer versions of the software. ML transparency mechanisms should ideally be expanded to include such comprehensive test artifacts.

% It is important to dedicate a system in place for assessing the  transparency and accountability of the ML systems and their testing protocol. In \cite{raji2020closing}, Raji {\it et al.}\ suggested an end to end framework to help with auditing an ML system. Datasheet \cite{gebru2018datasheets} and model card \cite{mitchell2019model} are practices driven from hardware community documentation practices or nutrition labeling. They mean to improve transparency, and accountability in ML systems. These practices should be extended in testing documentations, transparency and accountability.  

\textit{Recommendations:}

\begin{itemize}
    \item Consider expanding ML transparency mechanisms such as model cards \cite{mitchell2019model} to include a more comprehensive set of test results.
    \item Consider documenting the processes that went into building the set of test cases, so that the consumer of the ML system has a better understanding of its reliability.
    \item Consider documenting internal/external auditing practices (e.g., \cite{raji2020closing}) for both for model outputs and processes.
\end{itemize}

\section{Conclusion}

In this paper, we presented an overview of methodological shortcomings in traditional ML model testing. While we recognize the importance of
testing against distribution shift, we argue in favor of going beyond the sole focus on distribution driven testing, and take into consideration edge cases and severity of failures while assessing an ML model's performance. We draw inspiration from decades of work within software engineering testing practices, focused on assessing a software system against various stress conditions, measuring severity of failure cases and assessing test priorities. Based on this, we recommend a comprehensive set of considerations around the conditions to test for, the process that guide the testing, and the artifacts produced as the result of the testing process.
{\small
\bibliographystyle{ieee_fullname}
\bibliography{NeurIPS_2021}
}
\appendix

\begin{comment}
\section{Appendix}

Optionally include extra information (complete proofs, additional experiments and plots) in the appendix.
This section will often be part of the supplemental material.

\BHcomment{I am trying here to think through how some common metrics are characterisable as expected values of a function over a domain...
The point I am exploring is maybe these are all "statistical tests" (since they all involve expectations, which is a descriptive statistic).
Another point I am trying to explore is that comparing pairs of these is instructive. E.g. comparing precision with recall is instructive.
Similarly, \textbf{comparing} the domains over which expectations are calculated is instructive, e.g. a) model performance on training data vs on held out test data 
provides a measure of overfiiting, b) sliced evaluations for fairness evals.
\begin{itemize}
\item Accuracy (binary classification): $Pr\{Y = \hat{Y}\} = E(1 - |Y-\hat{Y}|)$
\item Sliced accuracy: $Pr\{Y = \hat{Y} | A=a\} = E(1 - |Y-\hat{Y} |A=a)$
\item Precision:$Pr\{Y = \hat{Y}| \hat{Y}=1\}= E(\hat{Y} | \hat{Y}=1)$
\item Recall:$Pr\{Y = \hat{Y}| Y=1\}= E(\hat{Y} | Y=1)$
\item Error rate (binary classification): $Pr\{Y \neq \hat{Y}\} = E(|Y-\hat{Y}|)$
\item MSE (regression):$E\{(Y -\hat{Y})^2\}$
\item Inverse Density Weighting (pay more attention to rare cases): ...
\item Severity weighting (pay more attention to severe cases): 
\end{itemize}
}

\end{comment}

\end{document}